%% file: neurips_2020.tex
\documentclass{article}

    \usepackage[preprint,nonatbib]{neurips_2020}

\usepackage[utf8]{inputenc} % allow utf-8 input
\usepackage[T1]{fontenc}    % use 8-bit T1 fonts
\usepackage{hyperref}       % hyperlinks
\usepackage{url}            % simple URL typesetting
\usepackage{booktabs}       % professional-quality tables
\usepackage{amsfonts}       % blackboard math symbols
\usepackage{nicefrac}       % compact symbols for 1/2, etc.
\usepackage{microtype}      % microtypography
\usepackage{subcaption}

\usepackage{tabularx}
\usepackage{amsmath,amssymb,amsthm}
\usepackage{bbm}
\usepackage{graphicx}
\usepackage{listings}
\usepackage{upquote}
\usepackage{array}
\usepackage{multicol}
\usepackage{multirow}
\usepackage{url}
\usepackage{color}
\usepackage{wrapfig,lipsum,booktabs}
\usepackage{makecell}
\usepackage{changepage}
\usepackage{xcolor}

\newcommand{\inline}[1]{{\small $#1$}}

\input{math_commands.tex}

\title{Deep Transfer Learning with Ridge Regression}

\author{%
\inline{^\alpha}Shuai Tang \hspace{1cm} \inline{^{\alpha\beta}}Virginia R. de Sa \\
\inline{^\alpha}Department of Cognitive Science, \inline{^\beta}Hal\i c\i o\u{g}lu Data Science Institute \\
University of California, San Diego \\
\{shuaitang93,desa\}@ucsd.edu
}

\begin{document}

\maketitle

\begin{abstract}
The large amount of online data and vast array of computing resources enable current researchers in both industry and academia to employ the power of deep learning with neural networks. While deep models trained with massive amounts of data demonstrate promising generalisation ability on unseen data from relevant domains, the computational cost of finetuning gradually becomes a bottleneck in transfering the learning to new domains. We address this issue by leveraging the low-rank property of learnt feature vectors produced from deep neural networks (DNNs) with the closed-form solution provided in kernel ridge regression (KRR).  This frees transfer learning from finetuning and replaces it with an ensemble of linear systems with many fewer hyperparameters. Our method is successful on supervised and semi-supervised transfer learning tasks.
\end{abstract}

\section{Introduction}

Besides soaring high on tasks they are trained on, deep neural networks have also excelled on tasks where datasets are collected from similar domains. Prior work \cite{Yosinski2014HowTA} showed that filters/parameters learnt in DNNs pretrained on ImageNet generalise better with slight finetuning than those learnt from random initialisations. Since then, applications in Computer Vision have had  major breakthroughs by initialising DNNs with pretrained parameters and finetuning them to adapt to new tasks. Similarly, Natural Language Processing (NLP) welcomed its ``ImageNet Era'' with large and deep pretrained language models including Bert \cite{Devlin2018BERTPO}, and performance on downstream NLP tasks has achieved state-of-the-art on a daily basis by employing more data and deeper models during pretraining, and using smarter methods for finetuning.

These advances in transfer learning using pretrained DNNs and finetuning, however, come with a large computational cost. An essential step to boost the performance on a given new task is to finetune the pretrained DNN until it converges, which is computationally intense since these models tend to have hundreds of millions of parameters. An alternative approach is to freeze the parameters and treat the pretrained DNN model as a feature extractor which produces abstracted vector representations of data samples with the knowledge from pretraining, and then train a simple classifier on top of these extracted vectors. But as the parameters are not adapted to the new task, the latter approach provides inferior performance to finetuning.

We here propose a new way of augmenting the latter approach without finetuning the DNN.  Our approach is to take an accumulation of feature vectors produced at different individual layers which encode various different aspects of the data. Since feature vectors are highly correlated with each other, as they are generated from a single DNN, only a few of them are needed to make predictions. We adopt the alignment maximisation algorithm for combining kernels \cite{Cortes2012AlgorithmsFL}, in which we first find a convex combination of linear kernels constructed from individual layers that gives  maximal alignment with the target kernel constructed from one-hot encoding of the labels. Then, we take the ensemble of feature vectors of  layers selected by non-zero elements in the sparse combination,
and make predictions using kernel ridge regression (KRR).

\section{Related Work}

Transfer learning with classical machine learning methods has been studied for a couple of decades \cite{Pan2010ASO}, including boosting \cite{Dai2007BoostingFT}, support vector machines \cite{Muandet2013DomainGV}, ridge regression \cite{Cortes2011DomainAI}, etc. These methods benefit from the transparency of classical machine learning models, and universal function approximators including boosting and kernel methods with strong theoretical guarantees. However, it is not easy to incorporate structural priors into regularising the learning process, such as our knowledge about images and text. This information is crucial in advancing machine learning systems.

Neural networks are also universal function approximators \cite{Hornik1989MultilayerFN}, and learnt vectorised representations are generalisable across tasks, with recent advances in various architecture designs specifically for individual types of inputs,
including convolutional layers for image recognition \cite{LeCun1998GradientbasedLA}, recurrent layers \cite{Elman1990FindingSI,Hochreiter1997LongSM} and transformers \cite{Vaswani2017AttentionIA} for text processing, etc.
Recent research has demonstrated that deep models pretrained on large amounts of training data give decent performance on unseen data sampled from relevant domains \cite{Yosinski2014HowTA} by finetuning. With growing depth of networks, the cost for finetuning becomes non-negligible. Efforts in knowledge distillation from deep models to shallow ones \cite{Ba2014DoDN,Hinton2015DistillingTK} and to simple ones \cite{Frosst2017DistillingAN} showed that neural networks can be simplified after learning, although the learnt transferable features can be potentially detrimented during distillation.

Our approach takes the best of both worlds by using feature vectors produced from multiple layers of a pretrained neural network but without explicit finetuning, and makes predictions with KRRs on a downstream task. With help from low-rank approximations, our approach only requires passing the training data once through a neural network without backpropagation.

\section{Method}
The key concept is to apply KRR with a few layers of feature vectors produced from a pretrained  neural network to make predictions, classification in our case, on a downstream task. The notations include:
\inline{\mX\in\mathbb{R}^{N\times d}} is the data matrix with \inline{N} samples with each sample in \inline{d}-dimensional space, \inline{\mY\in\mathbb{R}^{N\times c}} is the corresponding labels with one-hot encoding,
\inline{\mX_l\in\mathbb{R}^{N\times d_l}} is the flattened feature vectors produced at the \inline{l}-th layer from a pretrained neural network,
\inline{\mS\in\mathbb{R}^{M\times N}} is the random projection matrix that meets the requirement of subspace embedding with \inline{M\ll N},
\inline{\mI_p \in\mathbb{R}^{p\times p}} is the identity matrix,
\inline{L} is the number of layers in a pretrained neural network, and
 \inline{\alpha} is the regularisation term in ridge regression.
Other notations will be introduced as needed.

\subsection{Low-rank Approximation at Individual Layers}
\label{lowrank-countsketch}
Flattened feature vectors generated from neural networks are generally high-dimensional and redundant, therefore, we adopt theoretical work \cite{udell2019big} showing that big data matrices are approximately low rank, and use random projections to obtain low-rank approximations of high-dimensional feature vectors with many fewer dimensions. Given that the Nystr{\"o}m method is well-studied in approximating large-scale kernel matrices \cite{gittens2016revisiting}, we follow the formula to approximate a linear kernel \inline{\mX_l\mX_l^\top} as
\begin{align}
  \mX_l\left(\mS\mX_l\right)^\top(\mS\mX_l\mX_l^\top\mS^\top)^\dag\left(\mS\mX_l\right)\mX_l^\top
\end{align}%
where \inline{\dag} is the pseudo-inverse of a square matrix. If \inline{M \ll d}, which is mostly the case for feature vectors generated from neural networks, and the eigendecomposition is written as \inline{(\mS\mX_l\mX_l^\top\mS^\top)^\dag=\mQ_l\Lambda_l\mQ_l^\top},
 then the low-rank approximation of \inline{\tilde{\mX}} can be obtained by \inline{\mX_l\left(\mS\mX_l\right)^\top\mQ_l\Lambda_l^{-0.5}} with each sample in at most \inline{M}-dimensional space.
 As we aim to conduct layer-wise low-rank approximations, it is preferrable to apply sparse random projections instead of dense ones.
 Therefore, we consider a stack of \inline{s} CountSketch \cite{Clarkson2013LowRA} to approximate the sparse Johnson-Lindenstrauss Transformation \cite{Woodruff2014SketchingAA}.

In CountSketch, the random projection matrix \inline{\mS} is considered as a hash table that uniformly hashes \inline{N} samples into \inline{M} buckets with a binary value randomly sampled from \inline{\{+1, -1\}}
 so there is no need to materialise \inline{\mS}. Successful applications of CountSketch including polynomial kernel approximation \cite{Pham2013FastAS} and large-scale regressions are due to its scalability with theoretical guarantees when few hash tables are used \cite{Woodruff2014SketchingAA}.
 Generally, larger \inline{s} leads to better approximations, yet the performance improvement becomes marginal.
Prior work \cite{Jagadeesan2019UnderstandingSJ} showed that \inline{s=4} empirically works on real-world datasets, thus, we set \inline{s=4}, and it drastically reduces the cost for low-rank approximations at individual layers. The time complexity of Nystr{\"o}m is \inline{\mathcal{O}((sN+M^2+NM)d_l+(M+N)M^2)}.

With limited GPU memory, producing feature vectors for a downstream task given a pretrained neural network is often done in batches of samples. CountSketch is also well-suited in this situation as, technically, the approximation can be done in only one forward pass of \inline{\mX}.

\subsection{Convex Combination of Features across Layers by Learning Kernel Alignment}
\label{learningkernelalignment}
Storing feature vectors at \inline{L} layers has memory complexity at most \inline{\mathcal{O}(NML)}, thus we aim to select only a few layers that give the maximum alignment with the target. Specificially, a vector \inline{\boldsymbol{\mu}=[\mu_1, \mu_2, ..., \mu_L]^\top} is optimised to maximise the following alignment \cite{Cortes2012AlgorithmsFL}:
\begin{align}
  \boldsymbol{\mu}^\star = \argmax_{\{||\boldsymbol{\mu}||=1,\boldsymbol{\mu}\geq 0\}}\frac{\langle \mK, \mY\mY^\top \rangle_F}{||\mK||_F}
  \text{, where }\mK = \textstyle\sum_{l=1}^L \mu_l \tilde{\mX}_l\tilde{\mX}_l^\top \label{alignment}
\end{align}%
Proposition 9 in \cite{Cortes2012AlgorithmsFL} showed that it is equivalent to the quadratic programming problem: \inline{  \vv^\star = \textstyle{\argmax_{\vv \geq 0}} \vv^\top\mM\vv - 2\vv^\top\va},
 where  \inline{\va_l=||\tilde{\mX}_l^\top\mY||_F^2} and \inline{\mM_{k,l} = ||\tilde{\mX}_k^\top\tilde{\mX}_l||_F^2},
then \inline{\boldsymbol{\mu}^\star = \vv^\star / ||\vv^\star||}.
 Intuitively,
Non-zero entries in {\small $\boldsymbol{\mu}^\star$} provide a weighted sparse combination of feature vectors from a few layers that gives the highest linear alignment with targets.
The time complexity is dominated by materialising {\small $\mM\in\mathbb{R}^{L\times L}$}, which is {\small $\mathcal{O}(L^2M^2N)$} at worst.

The kernel \inline{\mK} induces an embedding space which is a concatenation of feature vectors weighted by \inline{\mu^{1/2}_l}, then the optimisation problem in Eq. \ref{alignment} can be written in a weight-space perspective:
\begin{align}
  \boldsymbol{\mu}^\star = \argmax_{\{||\boldsymbol{\mu}||=1,\boldsymbol{\mu}\geq 0\}}
  \frac{||\mX_\phi^\top\mY||_F^2}{||\mX_\phi^\top\mX_\phi||_F} \text{, where } \mX_\phi = [\mu_1^\frac{1}{2}\tilde{\mX}_1, \mu_2^\frac{1}{2}\tilde{\mX}_2, ..., \mu_L^\frac{1}{2}\tilde{\mX}_L]
\end{align}%
It is worth noting that the objective is not the ``goodness-of-fit'' measure for linear regression,  \inline{R^2} statistics \inline{||\mX_\phi^\top\mQ_Y||_F^2/||\mX_\phi||_F^2}, where \inline{\mQ_Y} contains eigenvectors of \inline{\mY\mY^\top}. Optimising \inline{\boldsymbol{\mu}} to maximise \inline{R^2} will lead to drastic overfitting by accumulating all layers, and subsequently meaningless \inline{\boldsymbol{\mu}}. The aforementioned objective finds a convex combination of features that maximises the alignment between the subspace spanned by the concatenated features and that by the onehot encoded label space, so it prevents \inline{\mX_\phi} from accumulating more feature vectors once an optimal subset is obtained. Therefore, the alignment-based objective prevents overfitting to a certain degree.

\subsection{[Optional Step] Nystr{\"o}m for Large-scale Kernel Approximation}
\label{nystrom}
We denote \inline{L_s} as the number of layers with positive \inline{\mu_l}'s. Since, in the end, the predictions are made by kernel ridge regression, if \inline{N} is at a manageable order, then there is no need to conduct kernel approximation through Nystr{\"o}m. However, low-rank approximation can potentially help reduce the noise in data, which leads to a better generalisation compared to computing the exact kernel function.

We consider approximating an RBF kernel function \inline{k(\vx_i, \vx_j)=\exp(-||\vx_i-\vx_j||^2/2\sigma^2)} with the Nystr{\"o}m method using the same subsampling in Sec. \ref{lowrank-countsketch}, CountSketch, to further promote fast computation on accumulated feature vectors \inline{\mX_\phi}. We denote the number of buckets in \inline{m} hash functions as \inline{M_s}, then the time complexity of this step is \inline{\mathcal{O}((m+M_s)NM+2M_s^2M+M_s^3)}.
Since \inline{N\gg M} and \inline{N\gg M_s}, the dominating term in the complexity is \inline{M_sNM}. The hyperparameter \inline{\sigma^2} is heuristically set to \inline{\max\{||\vx_i||^2/2 : i = 1,2,...,N\}}. One could cross-validate \inline{\sigma^2} as well, however, for the sake of reducing of the complexity of transfer learning, we stick to the heuristic value.

\subsection{Ridge Regression for Predictions}
\label{ridgeregression}
The approximated low-rank feature map of an RBF function is denoted as \inline{\tilde{\mX}_\psi\in\mathbb{R}^{N\times M_s}}. Given a new data sample \inline{\tilde{\vx}_\psi}, the prediction is given the closed-form solution of ridge regression in the table.
\begin{table}[h]
  \centering
\begin{tabular}{ c|c|c }
 \toprule
  condition & \inline{N\gg M_s} & \inline{N\ll M_s} \\
  \midrule
  prediction \inline{\vy}
  & \inline{\tilde{\vx}_\psi(\tilde{\mX}_\psi^\top\tilde{\mX}_\psi+\alpha\mI_{M_s})^{-1}\tilde{\mX}_\psi^\top\mY}
  & \inline{\tilde{\vx}_\psi\tilde{\mX}_\psi^\top(\tilde{\mX}_\psi\tilde{\mX}_\psi^\top+\alpha\mI_{N})^{-1}\mY} \\
 \bottomrule
\end{tabular}
\end{table}%
Then the label of a test sample is the index of the maximum value in predicted \inline{vy}. The time complexity of ridge regression is determined by the inverse of a square matrix and the matrix multiplication that gives the square matrix, and it is \inline{\min(N^3+M_sN^2, M_s^3+NM_s^2)}.

In summary, our proposed method has four steps including 1) \textbf{CountSketch} to obtain low-rank feature vectors at individual layers to a manageable size, 2) \textbf{convex combination} to take weighted accumulation of feature vectors, 3) \textbf{Nystr{\"o}m} for approximating an RBF kernel, and 4) \textbf{KRR} to make predictions.
Compared to multiple forward and backward passes required in finetuning or training classifiers, our method drastically reduces the computational cost.

\section{Experiments}
We demonstrate the effectiveness of our method through experiments on transfering ResNet-based models \cite{He2015DeepRL,He2016IdentityMI} pretrained on the ImageNet dataset \cite{Deng2009ImageNetAL,Russakovsky2015ImageNetLS} to downstream tasks,
including three in-domain datasets, CIFAR-10, CIFAR-100 \cite{Krizhevsky2009LearningML}, STL10 \cite{Coates2011AnAO},
and three out-of-domain ones, Street View House Number (SVHN) \cite{Netzer2011ReadingDI}, Caltech-UCSD-200 (CUB200) \cite{WelinderEtal2010}, Kuzushiji49 \cite{Clanuwat2018DeepLF}\footnote{The full Kuzushiji49 dataset has 232k training images, whick takes too long to cross-validate hyperparameters for LogReg. Thus, the same half of the dataset is used.}  . Basic statistics of each dataset are presented in Table \ref{dataset-details}.
\begin{table}[t]
  \caption{\textbf{Dataset details.} Individual cell indicates (\# Training Samples / \# Test Samples / [\# Classes]).}
  \label{dataset-details}
  \vspace{0.1cm}
  \fontsize{8}{10}\selectfont
  \centering
\begin{tabular}{c|ccc|ccc}
  \toprule
          Training & \multicolumn{3}{c|}{In-domain Transfer} & \multicolumn{3}{c}{Out-of-domain Transfer} \\
          ImageNet        & CIFAR10     & CIFAR100     & STL10     & SVHN                & CUB200      & Kuzushiji49       \\
    \midrule
     1.2m / - [1000]           & 50k / 10k [10]        & 50k / 10k [100]         & 5k / 8k [10]       & 73k / 26k [10]                & 6k / 6k [200]         &     116k / 38k [49]   \\
\bottomrule
\end{tabular}
\end{table}

\textbf{Hyperparameter Settings}: We report results with \inline{M=\{512, 1024, 2048\}}, and \inline{M_s=2M}. ResNet-18 and ResNet-34 pretrained on ImageNet are selected as base models to transfer from. To reduce the memory cost, instead of hashing all layers, we only hash feature vectors from every residual block in a model as each block usually has two or three convolutional layers. The regularisation strength \inline{\alpha} is cross-validated on the training set of the downstream task with values ranging from \inline{\{1e-1, 1e-2, 1e-3, 1e-4\}}.

\textbf{Comparison Partner}: \textbf{Finetuning the top layer on each downstream task with softmax regression.} Models are finetuned for 30 epochs with Adam optimiser, and the learning rate decays by a factor of 2 every 10 epochs. Cross validation is conducted to optimise the following hyperparameters and their associated values: data augmentation=\{with, without\}, weight decay rate=\{\inline{1e-3},\inline{1e-4},\inline{1e-5}\}, initial learning rate=\{\inline{1e-3}, \inline{2.5e-4}\}. Note that finetuning with data augmentation tremendously increases the training time as the neural network needs to be kept during finetuning, while for others one can store feature vectors from the last layer prior to finetuning. Results are marked with \textbf{LogReg} in the following tables and figures.

\textbf{Trials}: Since our method involves random projections and comparison partners require initialisation, for fair comparison, we run each method five times with different random seeds, and each marker in each plot presents the mean of five trials along with a vertical bar indicating the standard deviation. It is noticeable that vertical bars are often invisible as hyperparameters of each method are cross-validated on the training set. The main results are presented in Tab. \ref{overall}.
\begin{table}[t]
  \caption{\textbf{Results of supervised transfer learning.} Median accuracy of five trials is reported in each cell, and each cell has two accuracy terms of transferring from [ResNet-18 / ResNet-34]. Expect for CUB200, our method outperforms LogReg significantly since the variance of five trials is very small as presented in figures. }
  \label{overall}
  \vspace{0.1cm}
  \fontsize{8}{10}\selectfont
  \centering
\begin{tabular}{lcccccc}
  \toprule
       & CIFAR10 {[}In{]} & CIFAR100 {[}In{]} & STL10 {[}In{]} & CUB200 {[}Out{]} & SVHN {[}Out{]} & Kuzushiji49 {[}Out{]} \\
       \midrule
LogReg & 87.45 / 89.94    & 69.08 / 72.76     & 95.08 / 96.55  & 60.80 / 61.60    & 64.36 / 59.47  & 74.56 / 71.08         \\
\midrule
Ours   & 90.77 / 92.31    & 71.31 / 74.63     & 96.30 / 97.31  & 58.78 / 61.70    & 88.76 / 88.53  & 88.12 / 88.00 \\
\bottomrule
\end{tabular}
\end{table}%

\subsection{Supervised Transfer with Varying Portions of Training Samples}
\begin{figure}[t]
    \centering
    \begin{subfigure}[b]{0.95\textwidth}
      \includegraphics[width=\textwidth]{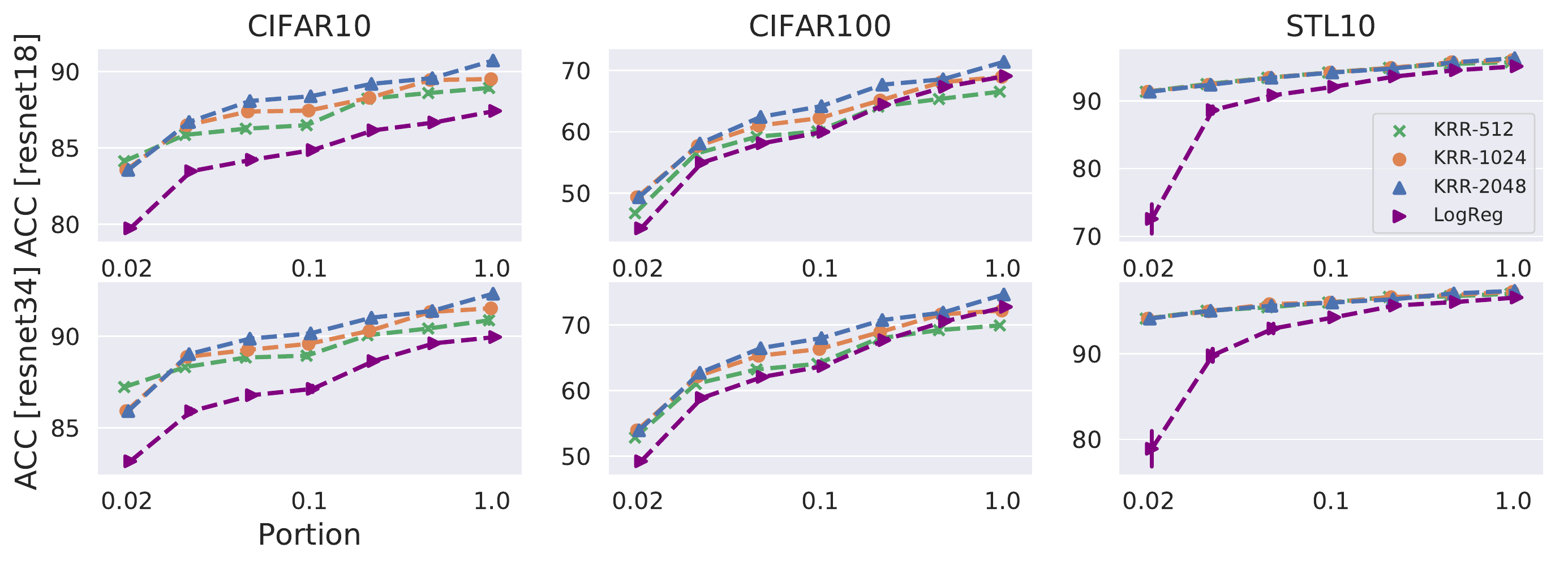}
      \caption{In-domain transfer tasks}
      \label{in-domain-acc}
    \end{subfigure}
    ~
    \begin{subfigure}[b]{0.95\textwidth}
      \includegraphics[width=\textwidth]{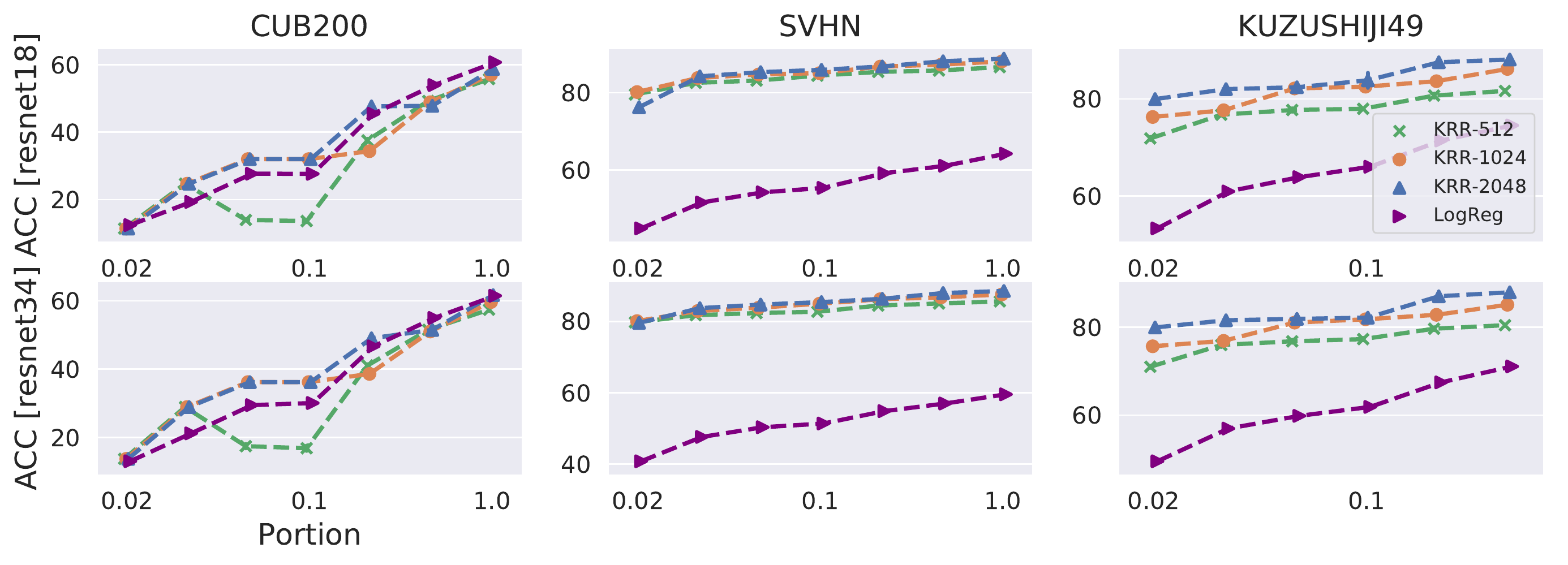}
      \caption{Out-of-domain transfer tasks}
      \label{out-of-domain-acc}
    \end{subfigure}
    \caption{\textbf{Supervised transfer with varying portions of training samples from the transfer task.} Except for CUB200, our method with all three \inline{M}'s generalises better than LogReg does (purple lines in plots) when the portion of training samples varies from \inline{2\%} to \inline{100\%}, and the observation is consistent across two different depths of ImageNet models.}
    \label{both-domain-acc}
\end{figure}

Since individual downstream tasks have ample samples in the training set, it encourages us to study our method and it comparison method when varying the portion of training samples. Specifically, the kept portion of training samples varies from \inline{2\%} to \inline{100\%}, and the interval is determined linearly in the log-space. The results are presented in Fig. \ref{both-domain-acc}.

Our method for \inline{M=\{1024, 2048\}} outperforms significantly finetuning the top layer on five out of six transfer tasks with different portions of training samples, and only performs relatively similar to finetuning on CUB200, which is a finegrained bird species recognition task.
\begin{figure}[t]
    \centering
    \includegraphics[width=\textwidth]{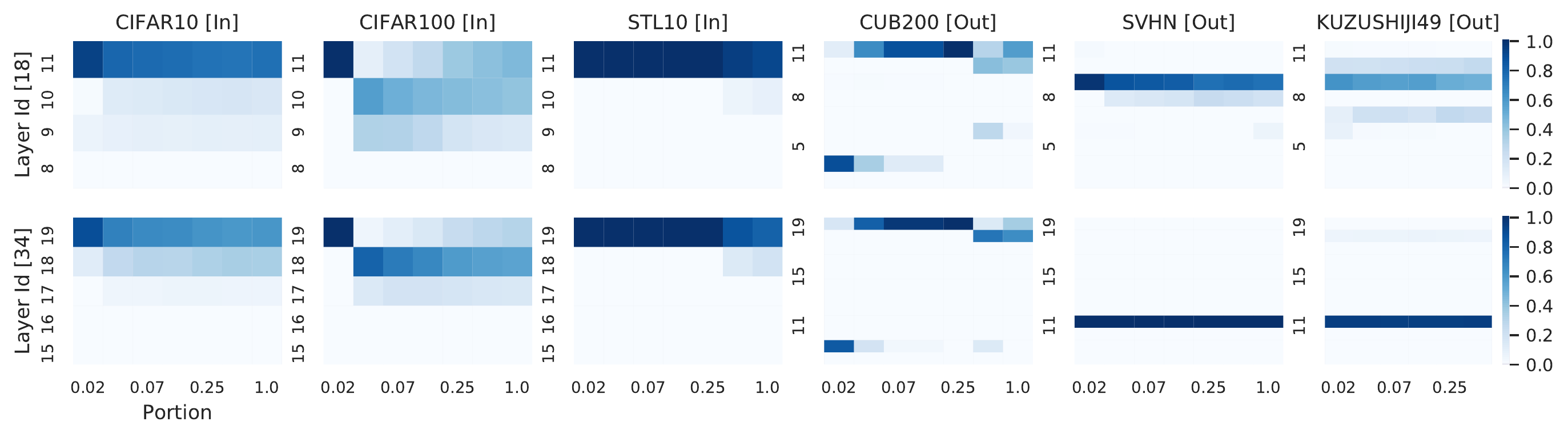}
    \caption{\textbf{Convex combination \inline{\boldsymbol{\mu}} of layers vs. Varying portions of training samples.} In-domain transfer tasks assign higher values to top few layers, and out-of-domain ones tend to give higher values to slightly lower layers.}
    \label{both-domain-mu}
\end{figure}%

\subsection{Insights provided by \inline{\mu_l}}
The solution to Eq. \ref{alignment} provides insights on the number of accumulated layers and their weights. We plot a heatmap with y-axis indicating the index of layers, x-axis indicating the portion of training samples, and gradient colour scheme presenting the value of \inline{\mu_l} in Fig. \ref{both-domain-mu}.

As shown in \cite{He2016IdentityMI}, the penultimate layer (index 11 for resnet18 and 19 for resnet34) of a ResNet removes all spatial information by averaging outputs from the previous layer.   As illustrated in Fig. \ref{both-domain-mu}, layers before the penultimate layer have been assigned non-zero \inline{\mu_l}'s across six tasks confirming that preserving spatial information helps in transfer learning.

For in-domain transfer tasks, it turns out that the top few layers are the  most useful, and the improvement of our method is brought by the ability of identifying and accumulating these layers. STL10 contains images from the ImageNet dataset  but with lower resolutions, so the penultimate layer provides adequately abstract information of the images, which explains the observation that our method assigns a very dominating \inline{\mu} towards the penultimate layer.

For SVHN and Kuzushiji49, clearly, the selected layers don't include the feature vectors generated from the penultimate layer, and lower layers give higher \inline{\mu}, which results in better performance than finetuning the top linear layer. However, our method doesn't provide better performance compared to finetuning the last layer on CUB200. A potential explanation comes from the fact that kernel ridge regression learns one-vs-all classifiers, and it is suitable when many classes are presented.
This is a limitation of our method, but also a research direction for future study.

\begin{figure}[t]
    \centering
    \begin{subfigure}[b]{0.95\textwidth}
      \includegraphics[width=\textwidth]{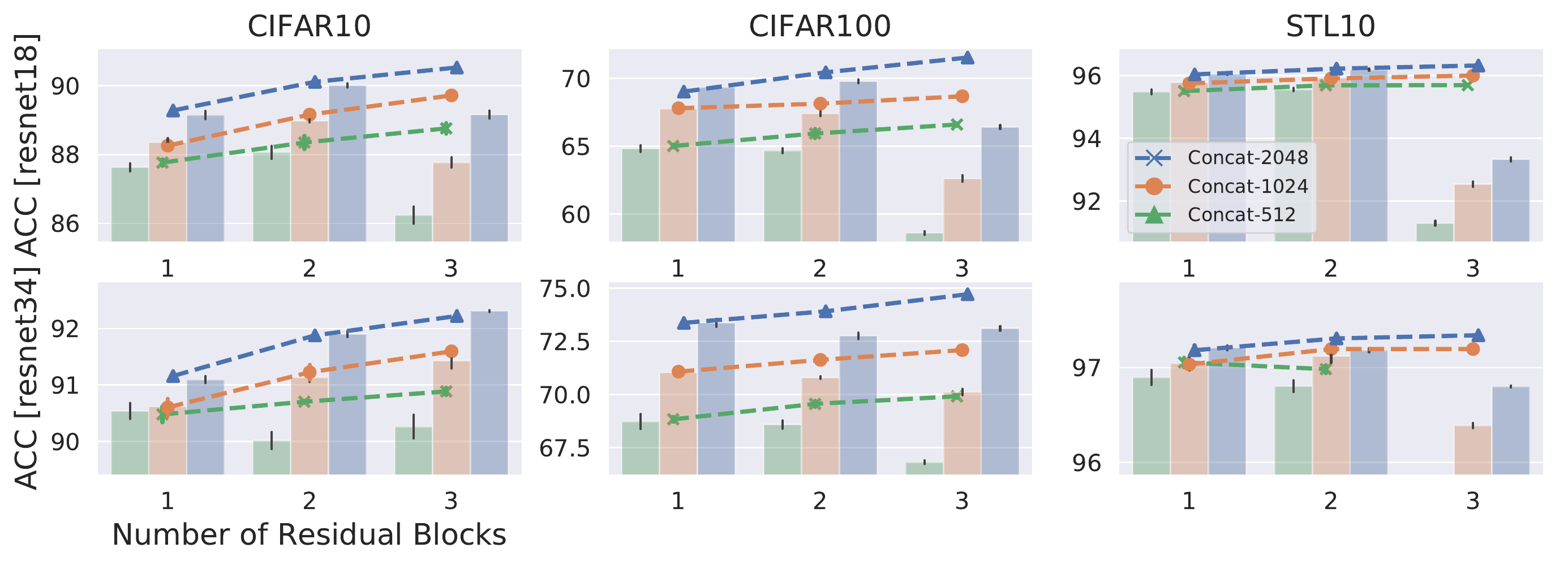}
      \caption{In-domain transfer tasks}
      \label{in-domain-AccumulatedVSIndividual}
    \end{subfigure}
    ~
    \begin{subfigure}[b]{0.95\textwidth}
      \includegraphics[width=\textwidth]{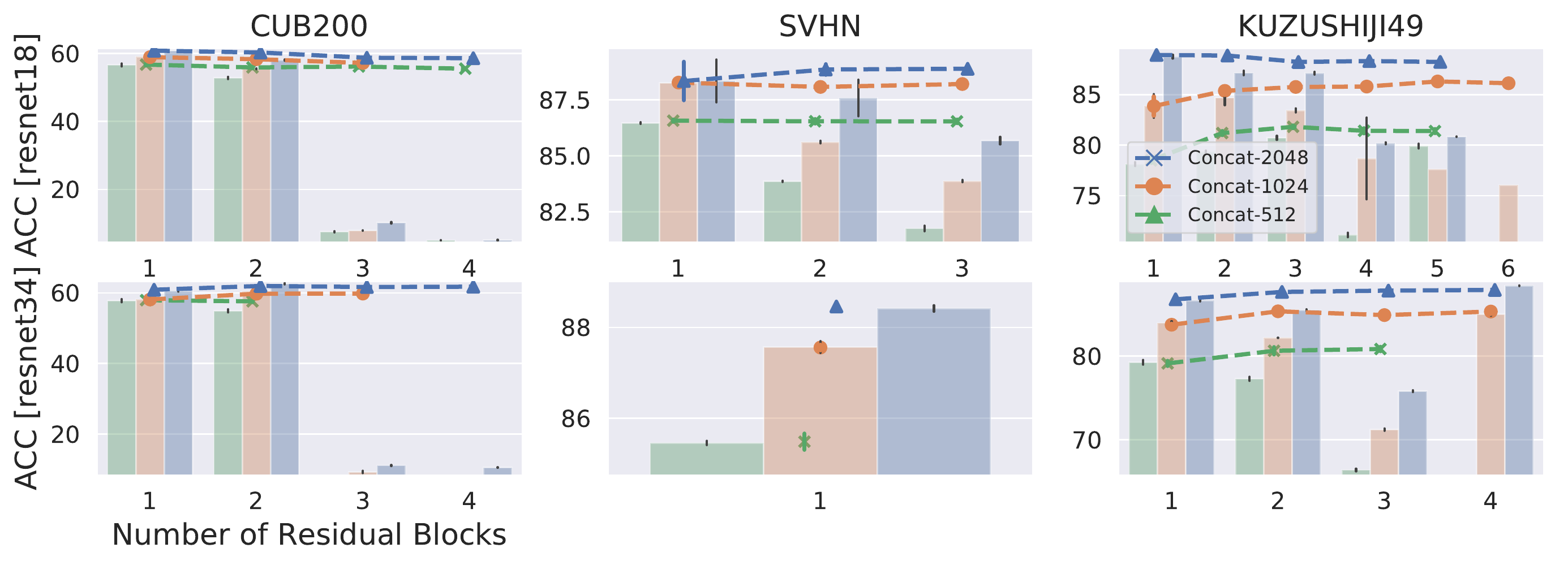}
      \caption{Out-of-domain transfer tasks}
      \label{out-of-domain-AccumulatedVSIndividual}
    \end{subfigure}
    \caption{\textbf{Accuracy of accumulating Residual Blocks vs. that of individual Ones.} Line plots indicate accuracy of accumulating blocks until the exhaustion of non-zero \inline{\mu_l}, and bar plots indicates the performance of these blocks separately. \textbf{(I)} In-domain transfer tasks demonstrate increasing accuracy when blocks are accumulated gradually, so does Kuzushiji49, which validates that accumulating layers helps. \textbf{(II)} Note that values \inline{\{\mu_l\}_{l=1}^L} don't directly imply the importance of layers, and that explains why the bar plots don't have a monotonic trend.}
    \label{both-domain-AccumulatedVSIndividual}
\end{figure}%
\subsection{Accumulated Feature Vectors vs Individual Feature Vectors}
As \inline{\mX_\phi} in our method is a weighted concatenation of feature vectors from layers with non-zero \inline{\mu_l}'s, it is important to conduct a sanity check on the effectiveness of accumulating layers compared to using these layers alone. Therefore, we gradually accumulate layers sorted by their \inline{\mu_l}'s, and plot  the performance curve versus the number of accumulated layers. Then these layers are applied individually to make predictions as a comparison. We use the full training dataset in this subsection. The results are shown in Fig. \ref{both-domain-AccumulatedVSIndividual}.

For in-domain transfer tasks, we see that the performance improves as our method accumulates layers,
while the trend is not obvious/significant for out-of-domain transfer tasks. Overall, accumulating a few layers provides better performance than making predictions based on individual layers.

\begin{figure}[t]
  \centering
  \begin{subfigure}[b]{0.95\textwidth}
    \includegraphics[width=\textwidth]{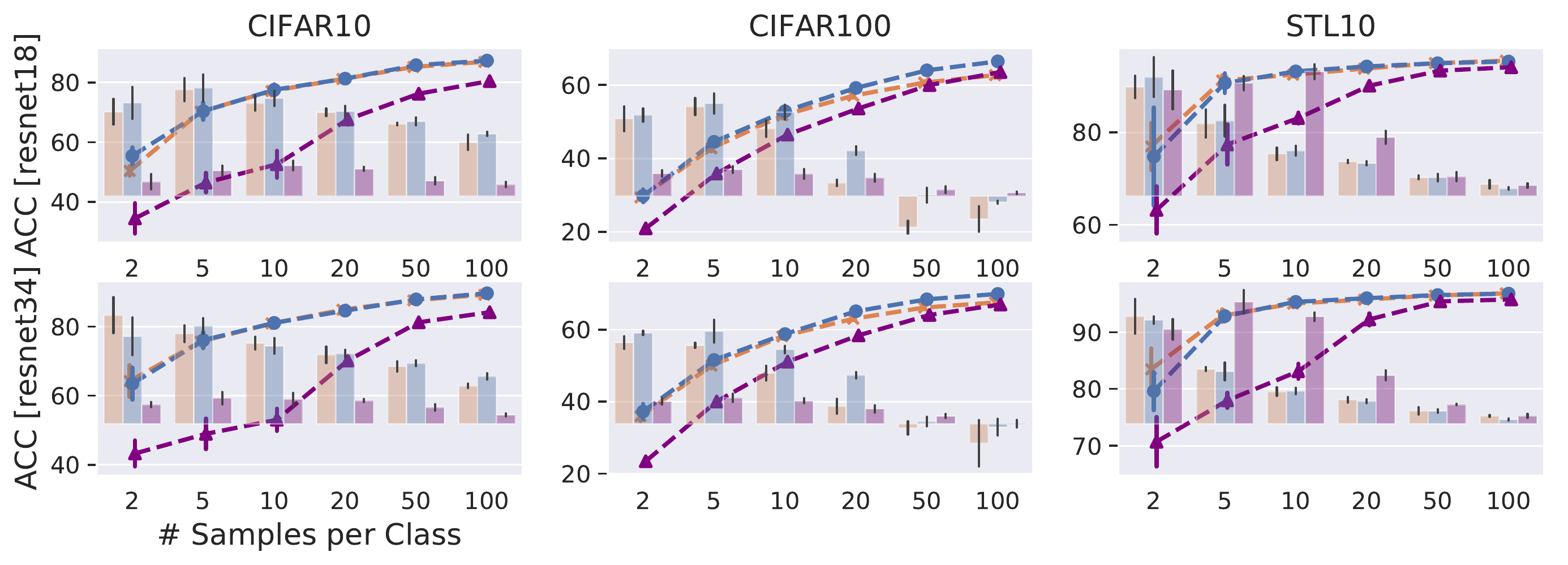}
    \caption{in-domain transfer tasks}
    \label{in-domain-semisupervised}
  \end{subfigure}
  ~
  \begin{subfigure}[b]{0.95\textwidth}
    \includegraphics[width=\textwidth]{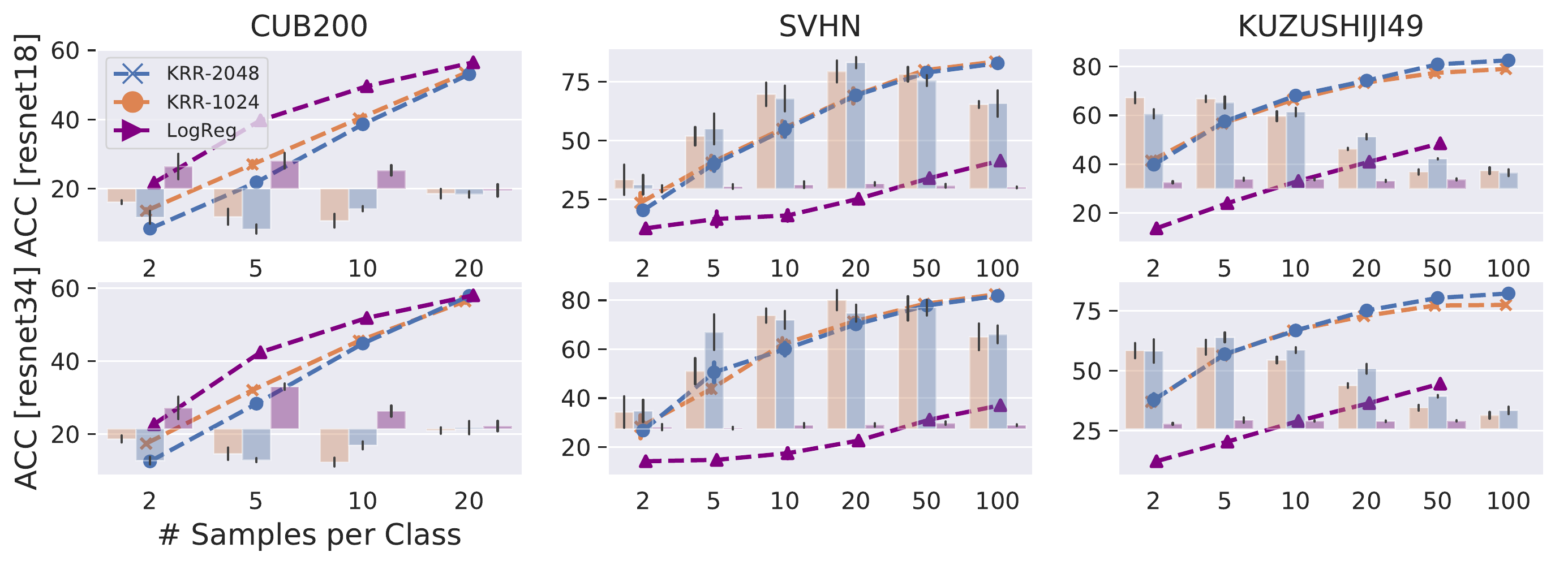}
    \caption{out-of-domain transfer tasks}
    \label{out-of-domain-semisupervised}
  \end{subfigure}
  \caption{\textbf{Accuracy of semi-supervised learning with varying number of labelled samples per class.} The number of labelled examples changes from \inline{1} to \inline{100} per class and the rest are left  unlabelled for semi-supervised learning. Left y-axis for line plots refers to the accuracy of semi-supervised learning, and right y-axis for bar plots refers to the relative improvement brought by unlabelled data.
  \textbf{(I)} Our method gives better performance than LogReg overall expect for CUB200.
  \textbf{(II)} Our method is also better at leveraging unlabelled samples for learning as indicated by taller bars for ours than LogReg expect for STL10 and CUB200.}
  \label{semisupervised}
\end{figure}%
\subsection{Semi-supervised Transfer Learning via Transductive Regression}
There are many ways of incorporating unlabelled data into kernel ridge regression, including manifold regularisation \cite{Belkin2006ManifoldRA} and transductive learning \cite{Cortes2006OnTR}. Since manifold regularisation requires exact computation or an approximation of the  Laplacian matrix on labelled and unlabelled samples, which leads to increased learning time,
we adopted the transductive learning method for regression problems to leverage unlabelled data when extremely limited labelled training samples \inline{\{\tilde{\mX}_\psi,\mY\}} with large amount of unlabelled samples \inline{\{\tilde{\mX}^\prime_\psi\}} are provided.
The solution of transductive ridge regression \cite{Cortes2006OnTR} is
\begin{align}
  \mW = (\beta^\prime\tilde{\mX}_\psi^{\prime\top}\tilde{\mX}_\psi^{\prime}
  +\beta\tilde{\mX}_\psi^\top\tilde{\mX}_\psi+\mI_{M_s})^{-1}
  (\beta^\prime\tilde{\mX}_\psi^{\prime\top}\mY^\prime+\beta\tilde{\mX}_\psi^\top\mY)
\end{align}
where \inline{\beta^\prime} and \inline{\beta} are hyparameters that control the contribution from unlabelled data and labelled data, which can be cross-validated on the labelled data. \inline{\mY^\prime} comes from the ridge regression model learnt only on labelled data, and is given as \inline{\mY^\prime=g(\tilde{\mX}_\psi^\prime(\tilde{\mX}_\psi^\top\tilde{\mX}_\psi+\alpha\mI_{M_s})^{-1}\tilde{\mX}_\psi^\top\mY)},
where {\small $\alpha$} is a hyperparameter, and {\small $g(\cdot)$} sets the maximum value of the vector prediction of a data sample to \inline{1} and the rest to \inline{0}. For finetuning the top layer, we use the supervised classifier trained on labelled samples to annotate unlabelled data samples, and incorporate these samples into the training set and retrain the classifier with cross-validation.

We simulate a semi-supervised learning environment by keeping 2, 5, 10, 20, 50, or 100 labeled training samples per class on each dataset, and leave the rest as unlabelled samples. The accuracy of semi-supervised transfer learning on the testset is reported in lineplots in Fig. \ref{semisupervised}, and the relative improvement against supervised transfer learning is reported in barplots in the same figure.

Our method outperforms finetuning the last layer on five out of six transfer tasks indicated by the lineplots. Our method also gives significant relative improvement when unlabelled samples are incorporated through transductive regression on these five tasks as well, while unlabelled samples don't improve the generalisation ability for the top layer finetuning method. Negative results are concentrated on CUB200, where unlabelled samples become detrimental to our method while helpful for finetuning the top layer.

\section{Discussion}
\begin{figure}[t]
  \centering
  \begin{subfigure}[b]{0.95\textwidth}
    \includegraphics[width=\textwidth]{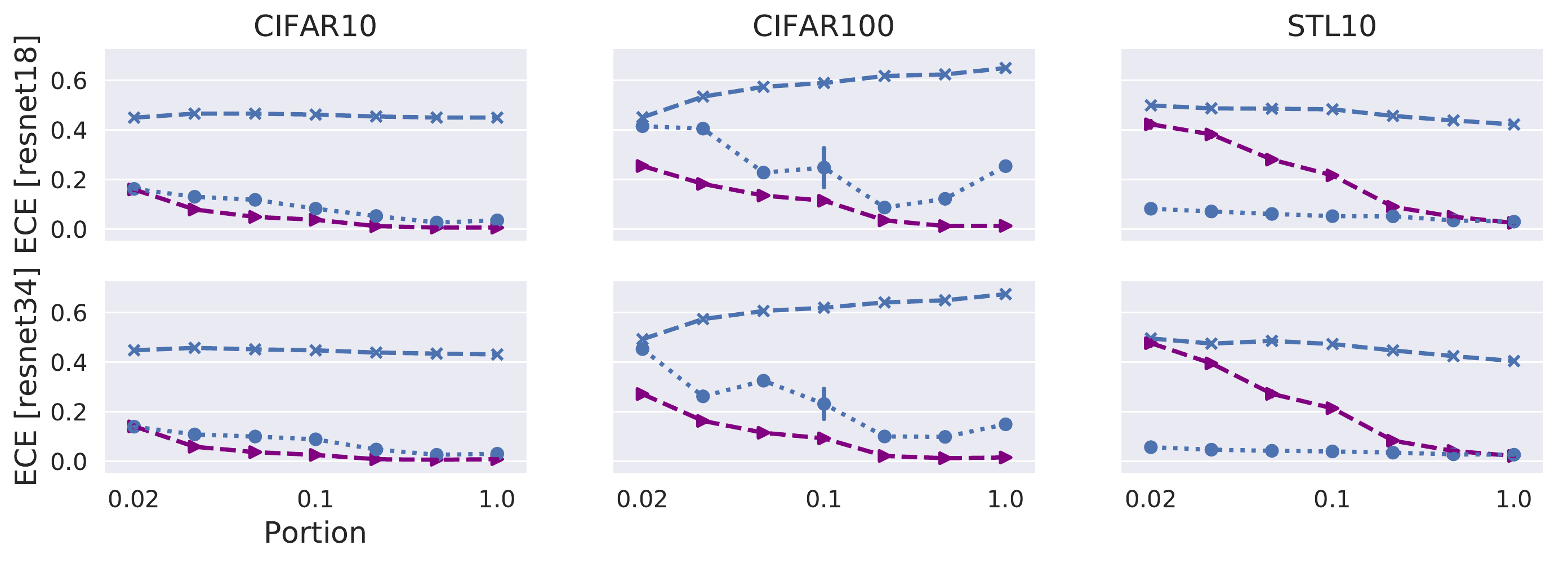}
    \caption{in-domain transfer tasks}
    \label{in-domain-eceloss}
  \end{subfigure}
  ~
  \begin{subfigure}[b]{0.95\textwidth}
    \includegraphics[width=\textwidth]{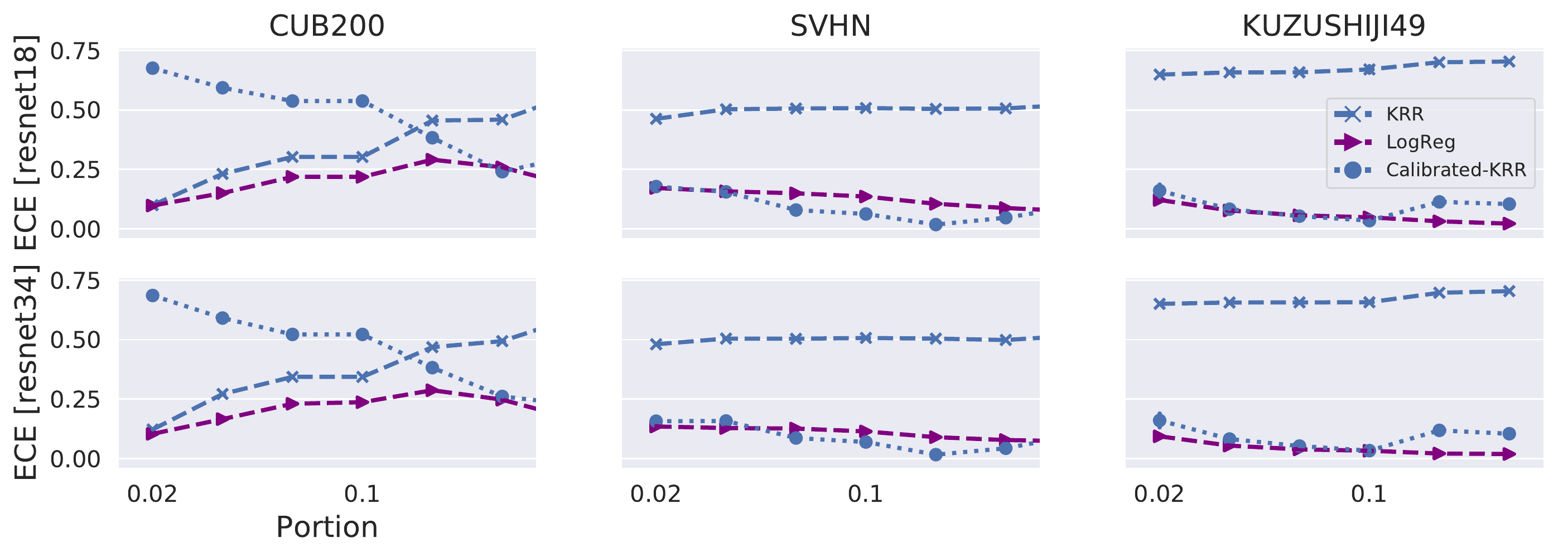}
    \caption{out-of-domain transfer tasks}
    \label{out-of-domain-eceloss}
  \end{subfigure}
  \caption{\textbf{Expected Calibration Error (lower the better) with varying portions of training data.} LogReg provides better calibrated models than ours does. However, a posthoc adjustment by \textit{Temperature Scaling} helps our method to match the calibration performance with LogReg. }
  \label{eceloss}
\end{figure}
\subsection{Temperature Scaling for Calibration}
Although our method provides both speed up and accuracy improvement in transfer learning, we are also interested in how well-calibrated our learnt classifier is compared to finetuning the top layer. It is expected that KRR, SVM and tree-based boosted classifiers are not well-calibrated as the predicted outputs can not be directly interpreted as classifers' confidence \cite{NiculescuMizil2005PredictingGP}. We calculate the Expected Calibration Error (ECE) \cite{Guo2017OnCO} for our method, and baseline models - finetuning the top layer with logistic regression.
The formula of ECE is given as \inline{\text{ECE}=\textstyle\sum_{c=1}^C\frac{B_c}{N}|\text{acc}(B_c)-\text{conf}(B_c)|},
where \inline{N} is the number of samples, and \inline{C} is the number of bins in the estimation.

The results shown in Fig. \ref{eceloss} validate our expectation that our method gives worse calibration on test set compared to logistic regression. A simple cure is Temperature Scaling \cite{Guo2017OnCO}, which optimises a parameter \inline{t} to rescale the output from the classifer in order to reduce the ECE on training data, and doesn't change the predicted labels. In our case, we can simply cross-validate \inline{t} very efficiently once the output from KRR is produced.

\begin{figure}[t]
  \centering
  \begin{subfigure}[b]{0.95\textwidth}
    \includegraphics[width=\textwidth]{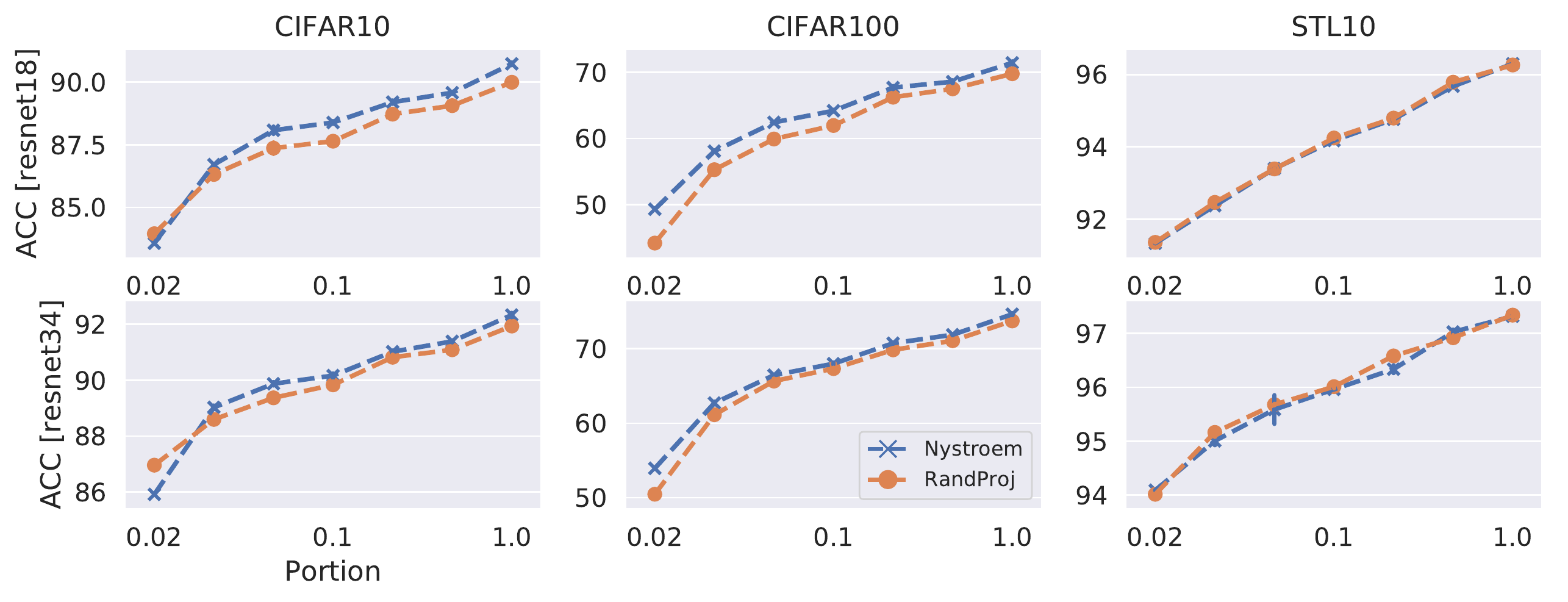}
    \caption{in-domain transfer tasks}
    \label{in-domain-nystroemvsrandproj}
  \end{subfigure}
  ~
  \begin{subfigure}[b]{0.95\textwidth}
    \includegraphics[width=\textwidth]{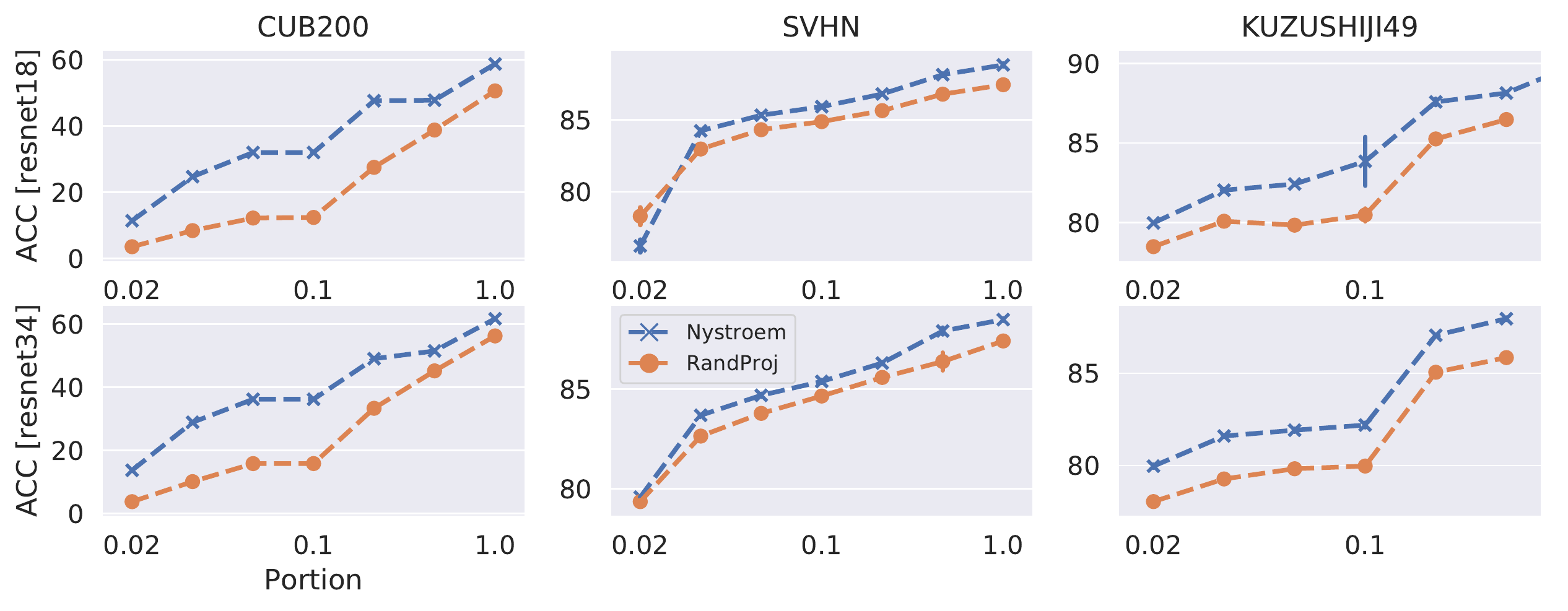}
    \caption{out-of-domain transfer tasks}
    \label{out-of-domain-nystroemvsrandproj}
  \end{subfigure}
  \caption{\textbf{Nystr{\"o}m vs. Random Projection in the first step of our method.} The performance improvement of Nystr{\"o}m over random projection is relatively larger on out-of-domain transfer tasks that it is on in-domain ones, and the observation is consistent across varying portions of training data.}
  \label{nystroemvsrandproj}
\end{figure}
\subsection{Random Projections}
Our method adopted the Nystr{\"o}m method for low-rank approximation of feature vectors at individual layers, which involves hashing \inline{N} data samples into \inline{M} buckets first, then solving a linear system, and the hash functions can be applied across all layers. A more direct approach is to hash individual features in each feature vector into \inline{M} buckets as in \inline{\mX_l\mS_l\in\mathbb{R}^{N\times M}}. This approach eliminates the step of solving a linear system, which reduces the time complexity to \inline{\mathcal{O}(sMd_l)} for each layer. The comparison between Nystr{\"o}m and random projection is presented in Fig.  \ref{nystroemvsrandproj}.

Overall, Nystr{\"o}m provides better accuracy across all six tasks than random projection. However, it is noticeable that the difference between the two is smaller on in-domain transfer tasks. The observation also serves as a piece of supporting evidence that our method is relatively consistent when different low-rank approximation schemes are applied in the first step.

\subsection{Task-dependent Distillation}
Our method in previous sections still requires the pretrained ImageNet model during testing time. Now we present task-dependent distillation to leverage the predictions constructed in our method as a regularisation in training smaller networks for individual downstream tasks.
\begin{wraptable}{l}{0.5\textwidth}
  \centering
  \caption{Predictions from our method as a regularisation for training small models on individual tasks. ``w/o'' refers to training models with cross-entropy loss only and ``w'' refers to training with MSE loss as a regulariser. The median performance of five random initialisations is reported. Overall, predictions from our method helps smaller models to generalise better.}
  \label{distillation}
  \fontsize{8}{9}\selectfont
  \begin{tabular}{cccc}
    \toprule
      & \multicolumn{3}{c}{ResNet 8 / ResNet 10}       \\
      & CIFAR10       & CIFAR100      & STL10         \\
      \midrule
  w/o & 91.50 / 92.22 & 69.24 / 69.41 & 67.31 / 69.21 \\
  w   & 92.23 / 93.22 & 69.03 / 70.87 & 69.91 / 71.64 \\
  \bottomrule
  \end{tabular}
  \vspace{-0.5cm}
\end{wraptable}
Once predictions on the training set of a task are made by KRR in our method, we store the predictions and remove the pretrained model. This step only requires memory complexity of \inline{\mathcal{O}(NC)}, where \inline{C} is the number of classes.

We train a ResNet with 8 layers and one with 10 layers on individual tasks with the cross-entropy loss, and use Mean Squared Error loss (MSE) to regress the output of neural networks to the predictions made by KRR from our method. The results are presented in Tab. \ref{distillation}. Overall, regularising small models with predictions of our methods from ImageNet models helps them to obtain better generalisation.

\section{Conclusion}
We provided a promising four-step Ridge Regression based transfer learning scheme for deep learning models. It doesn't require finetuning, which simplifies the transfer learning problem to simple regressions, and it is capable of identifying a few layers to accumulate for making better predictions.

We evaluated our method on supervised transfer with varying portions of training data, and handle semi-supervised transfer learning problems via transductive regression. Both show significant improvement compared to finetuning only the last layer.

Discussions addressed the issue of calibration by Temperature Scaling, and demonstrate the superiority of Nystr{\"o}m over plain random projections. Lastly, we showed that the predictions from our method can be used to improve shallow/small models when training directly on transfer tasks.

% \clearpage
% \input{main_paper/broaderimpacts.tex}
% \input{acknowledgements.tex}
\section*{Acknowledgements}

Shuai Tang and Virginia R. de Sa are supported by NSF IIS-1817226, and Shuai's research is partly funded by Adobe's gift funding. We gratefully thank Charlie Dickens and Wesley J. Maddox for fruitful discussions, and appreciate Mengting Wan and Shi Feng for comments on our draft.

\bibliographystyle{neurips_natbib}
\bibliography{example_paper}

% \clearpage
\appendix
\section{Dataset Descriptions}
\textbf{CIFAR10} \cite{Krizhevsky2009LearningML} consists of \inline{60k} images, each of size \inline{32 \times 32}. The train/test split is made available, and the training set contains \inline{50k} images and the test set contains the rest. Each image has an object at the center, and the total \inline{10} object categories are similar to ones in ImageNet dataset.

\textbf{CIFAR100} \cite{Krizhevsky2009LearningML} has \inline{60k} images as well, each of size \inline{32 \times 32}. The ratio of training images and test ones is the same as in CIFAR10. Each image has an object at the center, and in total, there are 100 object categories, which makes the task harder than CIFAR10.

\textbf{STL10} \cite{Coates2011AnAO} has \inline{500} images for training and \inline{800} images for testing per class, each of size \inline{96\times 96} and in total, there are \inline{10} classes. Since images of this dataset come from labelled samples from ImageNet but with lower resolution, models pretrained on ImageNet are expected to generalise well.

\textbf{SVHN} \cite{Netzer2011ReadingDI} consists of real-world images obtained from house numbers in Google Street View images, therefore, there are \inline{10} categories. The training set contains \inline{73,257} images, and the test set contains \inline{26,032}. The dataset also provides a set of \inline{531,131} unlabelled images, and we didn't make use of it in our study.

\textbf{CUB200} \cite{WelinderEtal2010} is an image dataset with photos of 200 bird species (mostly North American). The total number of training images is \inline{6,033}, therefore, each class has around 30 training examples. The task itself is considered to difficult as it requires the model to pay attention to details of the bird presented in each image, which makes it a fine-grained classification problem.

\textbf{Kuzushiji49} \cite{Clanuwat2018DeepLF} is a Japanese character recognition task, which contains \inline{48} Hiragana characters and one Hiragana iteration mark. The dataset itself is much larger than aforementioned ones, and contains only gray-scale images. The training set contains \inline{232,365} images, and in our study, we only used half of the whole set. The test set contains \inline{38,547} images, which is used to evaluate the effectiveness of our method and other methods.

\section{Results: RBF Baselines}

RBF kernels as universal kernels are widely used in many research domains.
Since we used features produced by neural network models learnt on the ImageNet dataset as inputs to an RBF kernel, it is reasonable to compare to the method that takes an ensemble of RBF kernels with various bandwidths and directly takes the vectorised images as inputs. Nystr{\"o}m approximation is applied to reduce the memory complexity.

Individual RBF kernels are selected as follows, and learning kernel alignment \cite{Cortes2012AlgorithmsFL} is also applied to find the optimal combination of RBF kernels with different bandwidths.
\begin{align}
   & k_{ij,p} = \exp( - ||\vx_i - \vx_j||^2 / (2^p\gamma)), \\
   \text{ where } & \gamma=\text{median}_{i,j\in \{1,2,...,N\}}||\vx_i - \vx_j||^2 \nonumber \\
  \text{ and } & p\in\{-2, -1, ..., 10\} \nonumber
\end{align}%
The results are presented in Tab. \ref{including-rbf}. Since RBF kernels are directly operating on pixels of images without neural networks, the performance is worse than ours or finetuning the top layer (LogReg). It serves as an supporting evidence that inductive biases (prior knowledge) introduced by convolutional layers are important in image recognition tasks.

\begin{table}[h]
  \caption{Results of LogReg, our method and RBF kernels.}
  \label{including-rbf}
  \centering
\begin{tabular}{l | cccccc}
  \toprule
Methods       & CIFAR10 {[}In{]} & CIFAR100 {[}In{]} & STL10 {[}In{]}  \\
       \midrule
LogReg & 87.45 / 89.94    & 69.08 / 72.76     & 95.08 / 96.55   \\
RBF    & 51.40    & 21.76     & 43.58   \\
\midrule
Ours   & 90.77 / 92.31    & 71.31 / 74.63     & 96.30 / 97.31   \\
\bottomrule
\end{tabular}
\end{table}

\section{Results: Transferring within In-domain tasks}
We have three in-domain transfer tasks, and train a model for each task then evaluate its performance on other tasks using our method. The results are presented in Tab. \ref{within-indomain}. Overall, our method provides reasonable performance across tasks, and it doesn't involve finetuning the models. Specifically, for STL10, as the dataset itself has very few images in the training set, models trained on CIFAR10 and CIFAR100 give better generalisation on STL10 than those trained on STL10 itself.

\begin{table}[h]
  \caption{Transferring within In-domain tasks.}
  \label{within-indomain}
  \centering
\begin{tabular}{c|ccc}
  \toprule
\multirow{2}{*}{Tasks for Pretraining} & \multicolumn{3}{c}{Transfer Tasks} \\
                                      & CIFAR10    & CIFAR100    & STL10   \\
                                      \midrule
CIFAR10                               & 94.61      & 57.16       & 82.81   \\
CIFAR100                              & 87.85      & 76.21       & 80.63   \\
STL10                                 & 73.72      & 43.65       & 67.85  \\
\bottomrule
\end{tabular}
\end{table}

\end{document}

%% file: math_commands.tex
\usepackage{amsmath,amsfonts,bm}

\def\eqref#1{equation~\ref{#1}}
\def\1{\bm{1}}

\def\va{{\bm{a}}}

\def\vv{{\bm{v}}}

\def\vx{{\bm{x}}}
\def\vy{{\bm{y}}}

\def\mI{{\bm{I}}}

\def\mK{{\bm{K}}}

\def\mM{{\bm{M}}}

\def\mQ{{\bm{Q}}}

\def\mS{{\bm{S}}}

\def\mW{{\bm{W}}}
\def\mX{{\bm{X}}}
\def\mY{{\bm{Y}}}

\DeclareMathAlphabet{\mathsfit}{\encodingdefault}{\sfdefault}{m}{sl}
\SetMathAlphabet{\mathsfit}{bold}{\encodingdefault}{\sfdefault}{bx}{n}

\DeclareMathOperator*{\argmax}{arg\,max}